\documentclass[conference]{IEEEtran}
\IEEEoverridecommandlockouts

\usepackage{cite}
\usepackage{amsmath,amssymb,amsfonts}
\usepackage{algorithm}
\usepackage{algpseudocode}
\usepackage{graphicx}
\usepackage{textcomp}
\usepackage{xcolor}
\usepackage{booktabs}
\usepackage{siunitx}
\usepackage{multirow}
\usepackage{array}
\usepackage{amssymb}

\usepackage[hidelinks,colorlinks=true,allcolors=blue]{hyperref}

\def\BibTeX{{\rm B\kern-.05em{\sc i\kern-.025em b}\kern-.08em
    T\kern-.1667em\lower.7ex\hbox{E}\kern-.125emX}}
\begin{document}

\title{AMUSD: \underline{A}synchronous \underline{Mu}lti-Device \underline{S}peculative \underline{D}ecoding for LLM Acceleration}

\author{\IEEEauthorblockN{Bradley McDanel}
\IEEEauthorblockA{\textit{Department of Computer Science} \\
\textit{Franklin and Marshall College}\\
Lancaster, PA, USA \\
bmcdanel@fandm.edu}
}

\maketitle

\begin{abstract}

Large language models typically generate tokens autoregressively, using each token as input for the next. Recent work on Speculative Decoding has sought to accelerate this process by employing a smaller, faster draft model to more quickly generate candidate tokens. These candidates are then verified in parallel by the larger (original) verify model, resulting in overall speedup compared to using the larger model by itself in an autoregressive fashion. In this work, we introduce AMUSD (Asynchronous Multi-device Speculative Decoding), a system that further accelerates generation by decoupling the draft and verify phases into a continuous, asynchronous approach. Unlike conventional speculative decoding, where only one model (draft or verify) performs token generation at a time, AMUSD enables both models to perform predictions independently on separate devices (e.g., GPUs). We evaluate our approach over multiple datasets and show that AMUSD achieves an average 29\% improvement over speculative decoding and up to 1.96$\times$ speedup over conventional autoregressive decoding, while achieving identical output quality. Our system is open-source and available at \href{https://github.com/BradMcDanel/AMUSD/}{https://github.com/BradMcDanel/AMUSD/}.
\end{abstract}


\section{Introduction}
Large language models (LLMs) have revolutionized natural language processing, demonstrating remarkable capabilities across a wide range of tasks. These models, typically built on the Transformer~\cite{vaswani2017attention} architecture, generate text in an autoregressive manner, where each token is produced based on the previous tokens. While this approach yields high-quality outputs, it can be computationally intensive and time-consuming, especially for longer sequences. To address this challenge, researchers have developed various techniques to accelerate the generation process. One promising method is Speculative Decoding~\cite{stern2018blockwise}, which leverages a smaller, faster ``draft'' model to predict multiple tokens in advance. These candidate tokens are then verified by the original, larger model in parallel, potentially leading to significant speed improvements.

\begin{figure}
    \centering
    \includegraphics[width=\linewidth]{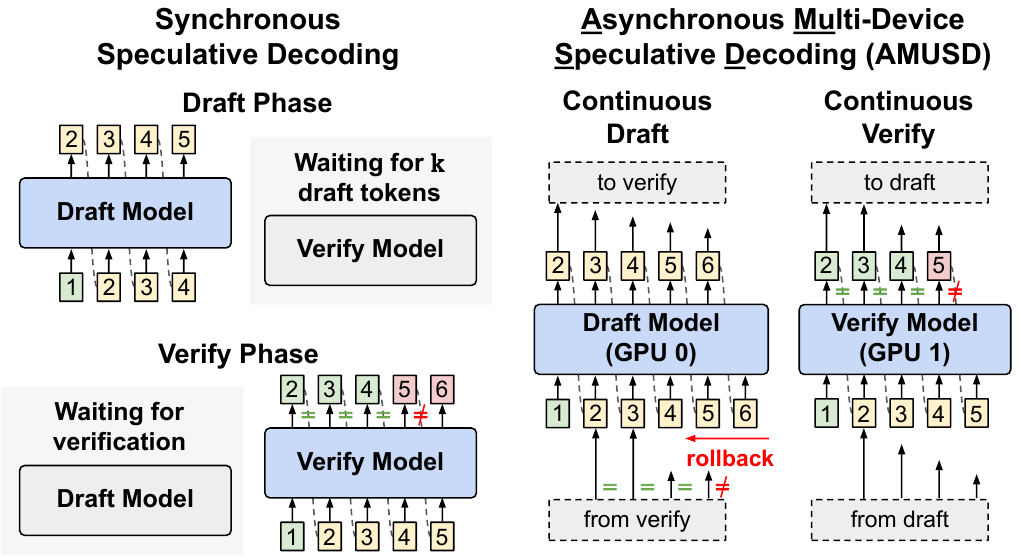}
    \caption{Comparison with synchronous speculative decoding (left) with AMUSD (right). The synchronous approach alternatives between drafting and verifying phases, meaning only model can work at a time. AMUSD uses asynchronous generation running on multiple GPUs enable continuous generation. The draft model must rollback invalid tokens that conflict with the verify model.}
    \label{fig:overview}
\end{figure}

While Speculative Decoding has shown promise in accelerating language model inference, it still faces limitations due to its synchronous nature. As illustrated in Figure \ref{fig:overview} (left), traditional speculative decoding alternates between distinct drafting and verifying phases, allowing only one model to perform computations at any given time. To overcome this constraint, we propose Asynchronous Multi-process Speculative Decoding (AMUSD), a novel system that decouples these phases into continuous, parallel operations. As shown in Figure \ref{fig:overview} (right), AMUSD leverages multiple devices (e.g., GPUs) to enable simultaneous and independent predictions from both the draft and verify models. This asynchronous approach allows for continuous generation, with the draft model producing candidate tokens while the verify model concurrently validates previously generated sequences. When conflicts arise between the draft and verify models, AMUSD implements a rollback mechanism to ensure output consistency. By leveraging parallel execution across multiple devices, AMUSD achieves substantial speed improvements over conventional speculative decoding methods, while producing identical generated text.

Our main contributions are as follows:
\begin{itemize}
    \item We introduce AMUSD, a novel asynchronous approach to speculative decoding that enables continuous parallel execution of draft and verify models on separate devices.
    
    \item We develop an efficient rollback mechanism that maintains output consistency while allowing for asynchronous operation.
    
    \item We implement and evaluate a complete system achieving up to 1.96$\times$ speedup over autoregressive decoding across diverse tasks.
\end{itemize}

\section{Related Work}
Speculative decoding has emerged as a promising approach for accelerating large language model inference. The core idea, first proposed by \cite{stern2018blockwise}, is to use a faster draft model to predict multiple tokens in advance, which are then verified in parallel by the original model. The field has since witnessed significant advancements \cite{xia2024unlocking}, from the development of multiple draft models and tree-structured verification \cite{miao2024specinfer} to the introduction of the MEDUSA framework \cite{cai2024medusa} with its novel multiple decoding heads approach. The ecosystem continues to evolve with optimizations such as recurrent mechanisms, token tree verification, distillation, and layer skipping techniques \cite{li2024eagle,he2024rest,zhoudistillspec,zhang2023draft,fu2024break}, enhancing both the efficiency and generalization capabilities of speculative decoding across various LLMs and tasks. Notably, these algorithmic innovations are orthogonal to the systems-level parallelization we propose, suggesting potential for combining these approaches to achieve even greater speedups.

Existing speculative decoding implementations typically operate in a synchronous manner, alternating between draft and verify phases. Our work, AMUSD, introduces a novel asynchronous approach that enables continuous parallel operation of both the draft and verify models. By decoupling the draft and verify phases, AMUSD achieves higher GPU utilization and lower latency compared to traditional speculative decoding methods. This advancement in efficient language model inference has potential applications across various natural language processing tasks, including code generation and software engineering, areas where large language models are increasingly being applied \cite{hou2023survey, zhou2023software, roziere2023code,jiang2023self}.

\section{AMUSD: Algorithmic Design and System}
\label{sec:amusd}
This section first describes the underlying principles that enable efficient speculative decoding using multiple devices working in an asynchronous fashion (Section~\ref{sec:amusd:overview}). Then, in Section~\ref{sec:amusd:alg}, we provide algorithms for the modifications required by asynchronous execution. Finally, we provide a detailed system implementation in Section~\ref{sec:amusd:system}.

\begin{algorithm}
\caption{Asynchronous Speculative Decoding}
\label{alg:amusd}
\begin{algorithmic}[1]
\Require Draft model $M_d$, Verify model $M_v$, Input $x$, Max tokens $N$
\State Initialize $D, V \leftarrow [\,]$, $p_d, p_v \leftarrow |x|$, $R \leftarrow \text{false}$
\State Initialize key-value caches $S_d, S_v$ with $x$

\Procedure{DraftProcess}{}
    \While{not finished}
        \State $t_d \leftarrow \text{GenerateToken}(M_d, S_d)$
        \State Append $t_d$ to $D$ at position $p_d$
        \State Update $S_d$ with $t_d$; $p_d \leftarrow p_d + 1$
        
        \If{$R = \text{true}$}
            \State Roll back $S_d$ to state at $p_v$
            \State $p_d \leftarrow p_v$; $R \leftarrow \text{false}$
        \EndIf
        
        \If{completion signaled} \textbf{break} \EndIf
    \EndWhile
\EndProcedure

\Procedure{VerifyProcess}{}
    \While{not finished}
        \If{$p_d > p_v$}
            \State $T_d \leftarrow D[p_v : p_d]$
            \State $T_v \leftarrow \text{VerifyTokens}(M_v, S_v, T_d)$
            
            \If{mismatch at position $i$ in $T_d$, $T_v$}
                \State Append $T_v[1 : i - 1]$ to $V$
                \State $p_v \leftarrow p_v + i - 1$; $R \leftarrow \text{true}$
            \Else
                \State Append $T_v$ to $V$; $p_v \leftarrow p_d$
            \EndIf
            
            \State Update $S_v$ with verified tokens
        \EndIf
        
        \If{end-of-sequence in $V$ or $|V| \geq N$}
            \State Signal completion; \textbf{break}
        \EndIf
        
        \If{$R = \text{true}$} \State Wait for draft rollback \EndIf
    \EndWhile
\EndProcedure

\State Run \textsc{DraftProcess} and \textsc{VerifyProcess} concurrently
\State \Return $V[1 : p_v]$
\end{algorithmic}
\end{algorithm}

\subsection{Overview}
\label{sec:amusd:overview}
\begin{figure*}
    \centering
    \begin{minipage}[b]{0.33\textwidth}
        \centering
        \includegraphics[width=\linewidth]{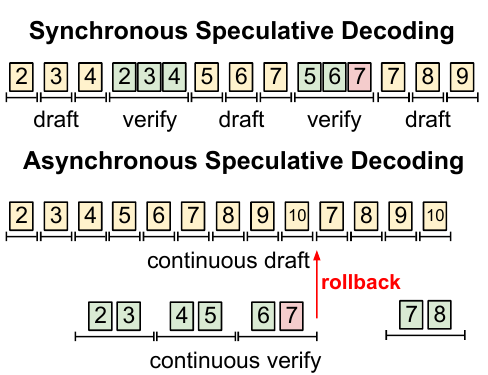}
        \caption{Synchronous decoding (top) alternates between draft and verify phases, while asynchronous decoding (bottom) runs both processes in parallel, with rollbacks on invalidation.}
        \label{fig:timing}
    \end{minipage}
    \hfill
    \begin{minipage}[b]{0.65\textwidth}
        \centering
        \includegraphics[width=\linewidth]{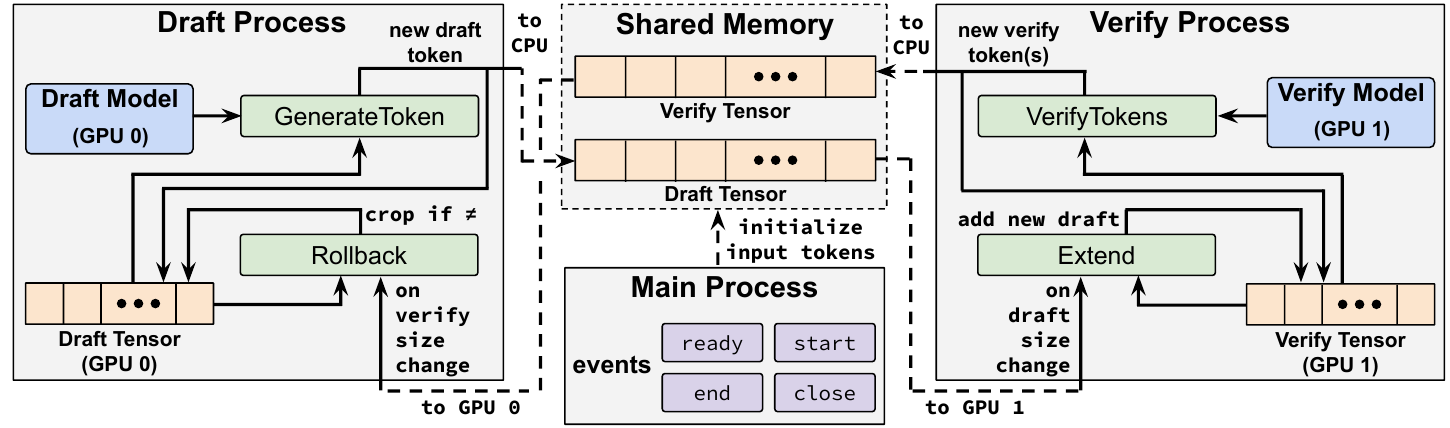}
        \caption{AMUSD system architecture showing the asynchronous interaction between Draft and Verify processes via shared memory. The Main Process coordinates execution while Draft (GPU 0) and Verify (GPU 1) processes maintain local tensors that sync with shared memory only when necessary, enabling efficient parallel execution while ensuring consistency through strategic updates and rollbacks.}
        \label{fig:system}
    \end{minipage}
\end{figure*}

Traditional speculative decoding runs on a single device, requiring draft and verify phases to strictly alternate. Our key innovation is enabling continuous, parallel execution by running these phases concurrently on separate devices. Figure~\ref{fig:timing} illustrates the difference in token generation between these synchronous and asynchronous approaches. In the synchronous case, after drafting tokens 2-4, the system must pause drafting to verify these tokens before proceeding. In contrast, our asynchronous approach continues drafting tokens 5-7 while verification of earlier tokens occurs in parallel. At the conclusion of the depicted sequence in the figure, the asynchronous method has verified tokens up to 8 and drafted up to token 10, while the synchronous method has only verified up to token 6 and drafted to token 9.

However, the asynchronous method introduces additional complexity in handling draft invalidations. Unlike the synchronous approach where these invalidations are resolved at predictable intervals after each verify phase, the asynchronous nature means that invalidations can occur at any time during the continuous drafting process. This necessitates a more dynamic rollback mechanism to ensure coherence between the draft and verify models and maintain output accuracy and consistency.

\subsection{Algorithmic Design}
\label{sec:amusd:alg}
Building upon the principles of asynchronous speculative decoding described in Section~\ref{sec:amusd:overview}, we now present the key algorithm that enables efficient implementation of AMUSD. Algorithm~\ref{alg:amusd} outlines the asynchronous speculative decoding process with the rollback mechanism.
The algorithm consists of two concurrent processes: drafting and verification. Both processes maintain shared access to draft and verify buffers ($D$ and $V$), along with their respective token lengths ($p_d$ and $p_v$) and model key-value caches ($S_d$ and $S_v$). A rollback flag $R$ coordinates the recovery process when token mismatches occur.
The draft process continuously generates tokens, updating its buffer and cache until either completion is signaled by the verify process or a rollback is required. Upon receiving a rollback signal, it reverts to the last verified position and resumes generation from there. The verify process monitors the draft buffer, validating new tokens in batches. When it detects a mismatch, it triggers the rollback mechanism by setting $R$ to true and waits for the draft process to complete its rollback before continuing.
This design allows for efficient parallel execution while maintaining consistency between the draft and verify models. The simplified initialization and streamlined coordination between processes reduces overhead while preserving the key benefits of asynchronous execution illustrated in Figure~\ref{fig:timing}. Note that locking of $D$ and $V$ are not required as each has only one writer (either the draft of verify process).

\subsection{System Architecture}
\label{sec:amusd:system}
Building upon Algorithm~\ref{alg:amusd}, our implementation physically separates the draft and verify processes across distinct devices (e.g., GPUs, CPUs, or specialized accelerators), with coordination managed through shared CPU memory, as illustrated in Figure~\ref{fig:system}. The main process orchestrates execution through control events implemented as \texttt{multiprocessing.Event}(s). Two shared tensors in CPU memory facilitate the asynchronous communication described in Section~\ref{sec:amusd:alg} - a draft tensor for newly generated draft tokens and a verify tensor for validated tokens.

As shown in the left and right portions of Figure~\ref{fig:system}, each process maintains a local tensor on its respective device that synchronizes with the shared CPU memory only when necessary. The draft process's local tensor expands with each new token generation and implements the rollback mechanism through tensor cropping, while the verify process's local tensor extends as new draft tokens are validated. This design enables efficient device utilization while keeping memory requirements manageable through strategic tensor synchronization, regardless of the specific hardware configuration chosen for deployment.

\section{Experimental Evaluation}
\label{sec:exp}

In this section, we evaluate AMUSD's performance across multiple dimensions. We first describe our experimental setup, including hardware configuration and model choices (Section~\ref{sec:exp:setup}). We then present a comprehensive performance analysis comparing AMUSD against baseline approaches across three diverse datasets designed to test different aspects of model generation (Section~\ref{sec:exp:perf}). Finally, we analyze resource utilization patterns and energy efficiency considerations across different decoding strategies (Section~\ref{sec:exp:resource}).

\subsection{Experimental Setup}
\label{sec:exp:setup}

All experiments were conducted on a system with 4 NVIDIA A100 GPUs and an AMD EPYC 7313 16-Core Processor. For AMUSD and the baseline speculative decoding implementation, we used Llama-3.1-8B~\cite{dubey2024llama} as the verify model and Llama-3.2-1B as the draft model. For a fair comparison, the autoregressive baseline used only the Llama-3.1-8B model. Note that the draft model takes $\sim$10 ms/token and the verify model takes $\sim$25 ms/token when run in a standalone autoregressive setting.

We evaluated performance across three diverse datasets designed to test different aspects of model generation. HumanEval~\cite{chen2021evaluating} provides 164 Python programming problems that test code generation capabilities through function implementation tasks with unit tests, covering algorithms, data structures, and mathematical operations. MT-Bench~\cite{zheng2023judging} contains 80 multi-turn dialogue scenarios designed to evaluate sophisticated conversational abilities including reasoning, roleplay, and writing tasks across diverse domains. To complement these established benchmarks, we introduce RefactorChat~\cite{mcdanel2024refactorchat}, a dataset of 100 extended software engineering interactions focused on code refactoring and feature additions. Each RefactorChat sample consists of 8 alternating user/assistant turns that simulate realistic development scenarios. For all datasets, we limited the response to 4096 tokens, but the majority completed early via an end of sequence token being emitted.

\subsection{Performance Analysis}
\label{sec:exp:perf}

Table~\ref{tab:results} presents the comparative performance across all three decoding strategies. AMUSD consistently outperforms both baseline approaches, with particularly notable gains on longer-sequence tasks. The improvements are most pronounced on RefactorChat, where AMUSD achieves a 1.96$\times$ speedup over autoregressive decoding and a 1.43$\times$ improvement over speculative decoding.

\begin{table}[t]
\centering
\caption{Performance comparison across decoding strategies.}
\label{tab:results}
\begin{tabular}{lccc}
\toprule
\textbf{Dataset} & \textbf{Method} & \textbf{Mean Token Time} (\(\downarrow\)) & \textbf{Speedup} (\(\uparrow\)) \\
\midrule
\multirow{3}{*}{HumanEval} & Autoregressive & 22.15 ms/token & 1.00$\times$ \\
& Speculative & 17.88 ms/token & 1.24$\times$ \\
& AMUSD (ours) & 14.08 ms/token & 1.57$\times$ \\
\midrule
\multirow{3}{*}{MT-Bench} & Autoregressive & 21.89 ms/token & 1.00$\times$ \\
& Speculative & 20.64 ms/token & 1.06$\times$ \\
& AMUSD (ours) & 16.75 ms/token & 1.31$\times$ \\
\midrule
\multirow{3}{*}{RefactorChat} & Autoregressive & 27.42 ms/token & 1.00$\times$ \\
& Speculative & 19.18 ms/token & 1.43$\times$ \\
& AMUSD (ours) & 13.96 ms/token & 1.96$\times$ \\
\bottomrule
\end{tabular}
\end{table}

\begin{figure}[t]
    \centering
    \includegraphics[width=\linewidth]{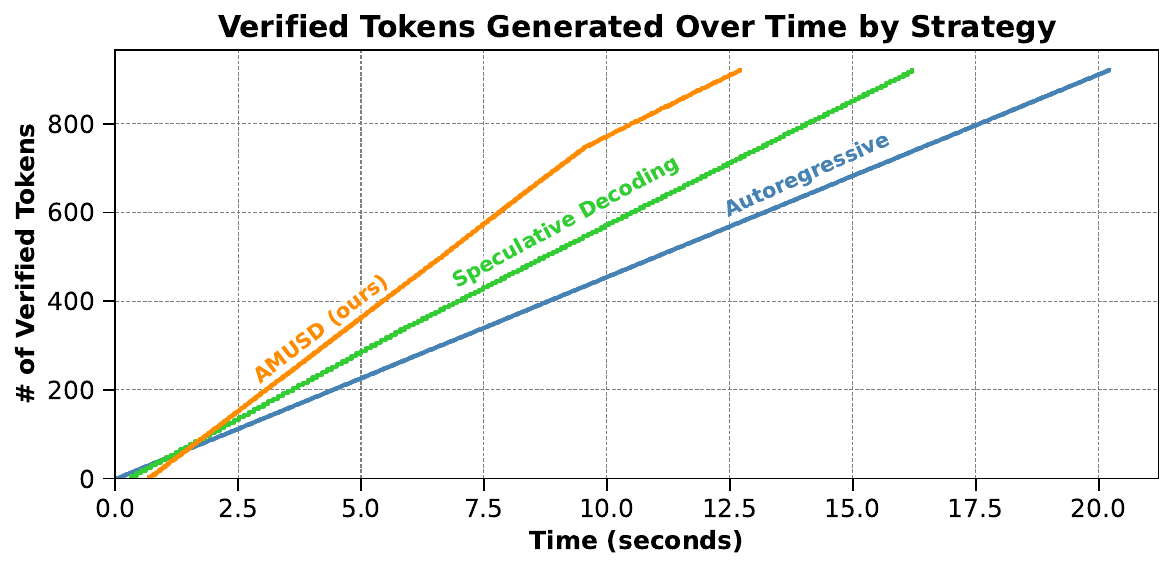}
    \caption{Token generation progress over time on a RefactorChat sample. AMUSD maintains consistently higher throughput due to parallel execution.}
    \label{fig:token-generation-comparison}
\end{figure}

Figure~\ref{fig:token-generation-comparison} visualizes token generation over time for a representative RefactorChat sample. The steeper slope of AMUSD's curve demonstrates how our asynchronous approach maintains consistently higher throughput throughout the generation process. This advantage stems from the continuous utilization of both models, as opposed to the alternating pattern in traditional speculative decoding. For this example, speculative decoding accepts 5.35 draft tokens per verify step compared to 2.75 draft tokens per verify step for AMUSD.

\subsection{Resource Utilization Analysis}
\label{sec:exp:resource}

\begin{figure}[t]
    \centering
    \includegraphics[width=\linewidth]{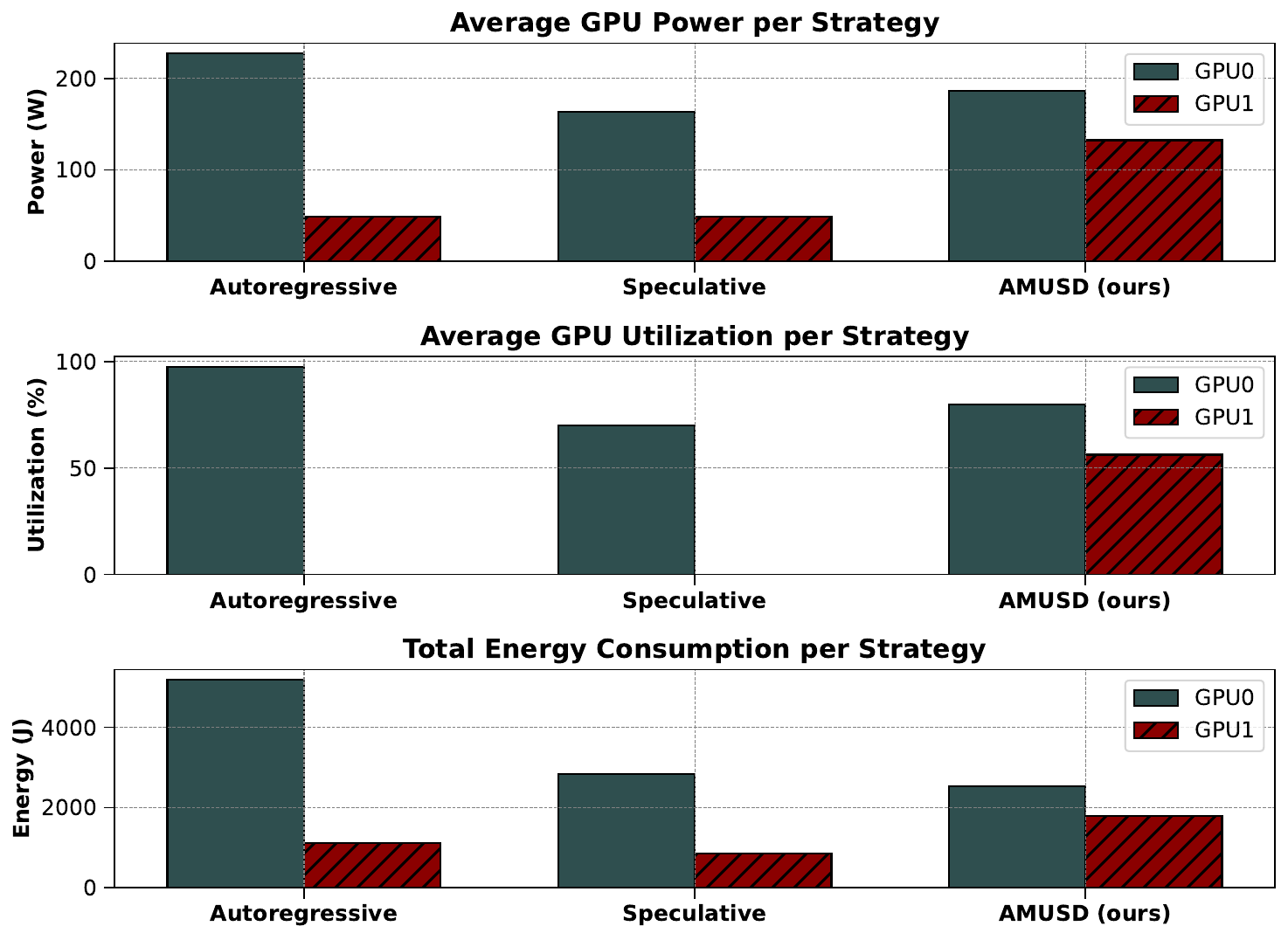}
    \caption{GPU resource utilization comparison across decoding strategies. Note that autoregressive and speculative decoding only actively use GPU0, with GPU1 shown to indicate baseline idle power consumption. Top: average power consumption. Middle: GPU utilization demonstrating resource use across devices. Bottom: total energy consumption per sample.}
    \label{fig:gpu-comparison}
\end{figure}

A key advantage of AMUSD is its efficient use of multiple GPUs through asynchronous execution. Figure~\ref{fig:gpu-comparison} compares GPU utilization patterns across the three approaches. While AMUSD shows higher total power consumption due to parallel model execution on two GPUs, it achieves significantly better overall throughput.

The results reveal an interesting trade-off in energy efficiency. While AMUSD requires approximately twice the instantaneous power due to concurrent model execution on two GPUs, its significantly faster completion times result in energy costs per token that are lower than autoregressive decoding and comparable to speculative decoding. For example, while AMUSD consumes around 300W across both GPUs compared to 150-220W for single-GPU methods, its smaller mean token time (see Table~\ref{tab:results}) leads to lower overall energy usage per token compared to autoregressive decoding and slightly higher energy usage compared to synchronous speculative decoding.

\section{Conclusion}
This paper introduced AMUSD (Asynchronous Multi-device Speculative Decoding), a novel system for accelerating large language model inference. By decoupling the draft and verify phases of speculative decoding into continuous, asynchronous operations across multiple GPUs, AMUSD achieves significant performance improvements over traditional methods. Our experiments demonstrate that AMUSD consistently outperforms both autoregressive and conventional speculative decoding across various benchmarks, achieving up to 1.96$\times$ speedup without compromising output quality. The system's ability to maximize GPU utilization through parallel, asynchronous processing represents a significant advancement in efficient LLM inference. As large language models continue to grow in size and importance, techniques like AMUSD will be crucial for making their deployment more practical and cost-effective across a wide range of applications.

\newpage
\bibliography{references}

\end{document}